\documentclass[letterpaper]{article} 
\usepackage{aaai24}  
\usepackage{times}  
\usepackage{helvet}  
\usepackage{courier}  
\usepackage[hyphens]{url}  
\usepackage{graphicx} 
\usepackage{subcaption}
\usepackage{natbib}  
\usepackage{caption} 
\usepackage{algorithm}
\usepackage{algorithmic}

\usepackage{amsmath, amsfonts, amsthm}
\newtheorem{theorem}{Theorem}

\usepackage{newfloat}
\usepackage{listings}
\DeclareCaptionStyle{ruled}{labelfont=normalfont,labelsep=colon,strut=off} 
\floatstyle{ruled}
\newfloat{listing}{tb}{lst}{}
\floatname{listing}{Listing}
\nocopyright
\title{SHLIME: Foiling adversarial attacks fooling SHAP and LIME}
\author{
    Sam Chauhan\textsuperscript{\rm 1},
    Estelle Duguet\textsuperscript{\rm 1},
    Karthik Ramakrishnan\textsuperscript{\rm 1},
    Hugh Van Deventer\textsuperscript{\rm 1},
    Jack Kruger\textsuperscript{\rm 1},
    Ranjan Subbaraman\textsuperscript{\rm 1}\\
}
\affiliations{
    \textsuperscript{\rm 1}University of Michigan\\
    \{csanjana, eduguet, kartikrk, hughv, 
    jakrug, ranjans\}@umich.edu
}

\begin{document}

\maketitle
\begin{abstract}
Post hoc explanation methods, such as LIME and SHAP, provide interpretable insights into black-box classifiers and are increasingly used to assess model biases and generalizability. However, these methods are vulnerable to adversarial manipulation, potentially concealing harmful biases. Building on the work of Slack et al. (2020), we investigate the susceptibility of LIME and SHAP to biased models and evaluate strategies for improving robustness. We first replicate the original COMPAS experiment to validate prior findings and establish a baseline. We then introduce a modular testing framework enabling systematic evaluation of augmented and ensemble explanation approaches across classifiers of varying performance. Using this framework, we assess multiple LIME–SHAP ensemble configurations on out-of-distribution models, comparing their resistance to bias concealment against the original methods. Our results identify configurations that substantially improve bias detection, highlighting their potential for enhancing transparency in the deployment of high-stakes machine learning systems.

\end{abstract}

\section{Introduction}


Our research problem focuses on the vulnerabilities of post hoc explanation methods like LIME (Local Interpretable Model-Agnostic Explanations) and SHAP (SHapley Additive exPlanations) to adversarial attacks. Post hoc means that these methods explain the decisions of a model after the model has already been trained and without changing how the model works. These explanation techniques are widely used in machine learning to provide interpretability of black-box models. It is important to understand the decision-making process of these black-box models as they are increasingly deployed in critical domains such as healthcare and criminal justice.

This project investigates how adversarial models can be constructed to deceive explanation methods, producing misleading or incorrect interpretations of model predictions. Specifically, it asks: How can adversarial attacks be designed to manipulate the outputs of LIME and SHAP while keeping the underlying model predictions unchanged? Such attacks are particularly concerning because the resulting biases are difficult to detect using post hoc explanation techniques.

 In this project, we will explore explainable AI (XAI), Adversarial Learning, and Model Interpretability and Robustness sub-areas in ML. In our replication, we will be using popular datasets for fairness in ML, including COMPAS, Communities and Crime, and German Credit. Both LIME and SHAP have readily available Python libraries with well-documented APIs that can be leveraged for this project. In addition, \cite{slack2020foolinglimeshapadversarial} has a GitHub repository that provides the necessary code to reproduce the project. This makes it easier to explore the adversarial attacks on post hoc explanation methods and reproduce the results of the experiments.

The project is computationally viable because the datasets mentioned above are small and can be processed efficiently. LIME generates local explanations efficiently, and while SHAP is more computationally expensive, it remains practical for small to medium datasets. The project focuses on devising adversarial attacks that involve manipulating input data rather than retraining entire models. This reduces computational complexity and makes the project both time-efficient and feasible within the available time frame.


\subsection{Related Work}
\textbf{The Mythos of Model Interpretability \cite{lipton2017mythosmodelinterpretability}:} This paper posits the idea that post hoc explanations do not guarantee that the model is unbiased. The paper argues that post hoc explanations can be optimized to disguise malicious intent in models, therefore using plausible-sounding explanations to mislead. \\
\\
\textbf{Interpretation of Neural Networks is Fragile \cite{ghorbani2018interpretationneuralnetworksfragile}:} This paper shows that even tiny, nearly imperceptible perturbations on the input can significantly alter the output of explainability tools. The paper also demonstrates that explanation techniques can be highly sensitive to small perturbations, even if the underlying classifier's predictions remain unchanged.\\
\\
\textbf{Explanations can be Manipulated and Geometry is to Blame \cite{dombrowski2019explanationsmanipulatedgeometryblame}:} This paper demonstrates how explanation methods are similarly vulnerable to imperceptible perturbations, resulting in a large change in output while the input remains visibly unchanged. The paper also proposes some mechanisms to improve the robustness of explanations in the context of visual models.

\section{Background}
In this section, we discuss the intuition and math underpinning seminal post hoc explanation techniques- LIME \cite{lime2016} and SHAP \cite{shap2017}.

\subsection{LIME and SHAP}
Complex models, such as ensemble models and deep neural networks, lack the natural interpretability found in simpler counterparts, such as linear algorithms and decision trees. These sophisticated models often operate as black boxes, making their decision-making process opaque. To gain insight into how these black boxes work, researchers can create simplified proxies that approximate and explain the behavior of complex models.

In this replication, we focus on linear explainer models. These models, known as additive feature attribution methods, are characterized by an explanation model that is a linear function of binary variables:

\begin{equation}
g(z') = \phi_0 + \sum_{i=1}^M \phi_i z_i'
\end{equation}

where $z' \in \{0,1\}^M$, $M$ is the number of simplified input features, and $\phi_i \in \mathbb{R}$.

Both LIME and SHAP seek to find interpretable models that locally approximate a complex model to explain how input features affect the complex model's output. They share a similar optimization framework:

\begin{equation}
\arg\min_{g \in \mathcal{G}} L(f, g, \pi_x) + \Omega(g)
\end{equation}

where
\begin{equation}
L(f, g, \pi_x) = \sum_{x' \in \mathcal{X}'} [f(x') - g(x')]^2 \pi_x(x')
\end{equation}

While LIME and SHAP share this fundamental structure, they differ in their specific implementations. LIME uses a distance-based weighting function $\pi_x(z) = \exp(-D(x,z)^2/\sigma^2)$ where $D$ is some distance function (such as L2 distance), defines $\Omega(g)$ as the number of non-zero weights in the linear model, and samples around the data point $x$ by adding noise.

SHAP, on the other hand, samples around $x$ by only extracting a random subset of input features, and grounds its selections for $\Omega$ and $\pi$ in game-theoretic principles, satisfying three key properties:
\begin{enumerate}
    \item Local accuracy (explanation must match original model output)
    \item Missingness (absent features must have zero impact)
    \item Consistency (increased feature contributions cannot decrease attribution)
\end{enumerate}

Notably, the authors demonstrate that Shapley values are the unique solution that satisfies these three axioms, providing a theoretical foundation for the SHAP approach to model interpretation. The authors then show the following selections for $\Omega$ and $\pi$ recover the Shapley values.

\begin{theorem}[Shapley kernel]
Under Definition 1, the specific forms of $\pi_{x'}$, $L$, and $\Omega$ that make solutions of Equation 2 consistent with Properties 1 through 3 are:

\begin{align*}
\Omega(g) &= 0, \\[1em]
\pi_{x'}(z') &= \frac{(M-1)}{\binom{M}{|z'|}|z'|(M-|z'|)}, \\[1em]
L(f,g,\pi_{x'}) &= \sum_{z' \in \mathcal{Z}} [f(h_x^{-1}(z')) - g(z')]^2 \pi_{x'}(z'),
\end{align*}

where $|z'|$ is the number of non-zero elements in $z'$.
\end{theorem}

\section{Proposed Framework for Adversarial Classifiers}

\subsection{Overview}
We discuss a novel adversarial framework to construct adversarial classifiers that can fool post hoc explanation techniques, such as LIME and SHAP \cite{slack2020foolinglimeshapadversarial}. These classifiers behave identically to a biased classifier \( f \) on real-world data but produce innocuous explanations when probed by explanation techniques. Below, we detail the framework's components: the setup and assumptions, the intuition behind the approach, and the methodology for constructing adversarial classifiers.

\subsection{Preliminaries}

\subsubsection{Problem Setting}
Consider an adversary seeking to deploy a biased classifier \( f \) to make critical decisions (e.g., parole, bail, or credit). Customers and regulators, with black-box access to \( f \), use post hoc explanation techniques to assess its fairness and readiness for deployment. If biases in \( f \) are detected, then the adversary will not be able to deploy the model. The adversary’s goal is to obscure \( f \)'s biases from these explanation techniques.

\subsubsection{Input}
The adversary provides the following input:
\begin{itemize}
    \item \textbf{Biased classifier} \( f \), intended for deployment.
    \item \textbf{Dataset} \( X \), sampled from the input distribution \( X_{\text{dist}} \), representing the real-world data on which \( f \) is applied.
\end{itemize}

\subsubsection{Output}
The framework outputs an \textbf{adversarial classifier} \( e \), which:
\begin{itemize}
    \item behaves identically to \( f \) on \( X_{\text{dist}} \).
    \item appears unbiased when analyzed by explanation techniques such as LIME or SHAP.
\end{itemize}

\subsection{Intuition Behind the Approach}
Post hoc explanation techniques, such as LIME and SHAP, approximate the behavior of a black-box model within the local neighborhood of a given data point by perturbing its input features. However, such perturbed instances often lie off-manifold or are out-of-distribution (OOD), making them distinct from real-world data points.

To demonstrate this, we performed principal component analysis (PCA) on a dataset combining original and perturbed instances generated by LIME. The visualization (Figure 1) revealed that the perturbed instances are clearly distinguishable from the original data. This insight suggests that distinguishing between real and perturbed instances is feasible, enabling the adversary to construct a classifier that behaves differently on perturbed data, thereby deceiving post hoc explanation techniques.

\begin{figure}[h]
    \centering
    \includegraphics[width=1\linewidth]{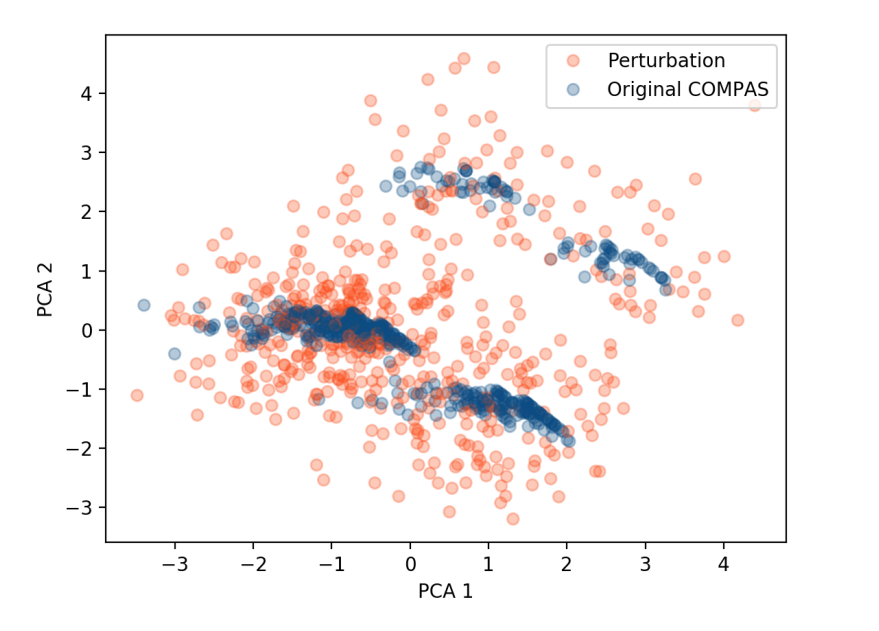}
    \caption{PCA applied to the COMPAS dataset (blue) as
well as its LIME style perturbations (red)}
    \label{fig:pca}
\end{figure}

\subsection{Framework Design}

\subsubsection{Adversarial Classifier Construction}
The adversarial classifier \( e \) is designed as follows:
\[
e(x) =
\begin{cases} 
f(x), & \text{if } x \in X_{\text{dist}} \\
\psi(x), & \text{otherwise}
\end{cases}
\]

Here:
\begin{itemize}
    \item \( f(x) \): Biased classifier behavior on real-world data.
    \item \( \psi(x) \): Unbiased classifier that makes predictions based on features that are uncorrelated with sensitive attributes.
\end{itemize}

This construction ensures that \( e \) mimics \( f \) on real data while exhibiting unbiased behavior on synthetic or perturbed instances.

\subsubsection{Detecting Out-of-Distribution (OOD) Samples}
To differentiate between real and perturbed data, we train a classifier \( \text{is\_OOD} \) as follows:
\begin{enumerate}
    \item Perturb each instance \( x \in X \) to generate a corresponding set \( X_p \) of perturbed instances.
    \item Assign labels:
    \begin{itemize}
        \item \( \text{False} \) for real data (\( X \)).
        \item \( \text{True} \) for perturbed instances (\( X_p \)), unless they overlap with \( X \).
    \end{itemize}
    \item Train a classifier on \( X \cup X_p \) using these labels.
\end{enumerate}

By accurately detecting OOD samples, \( \text{is\_OOD} \) enables the adversarial classifier \( e \) to behave selectively, presenting innocuous explanations for perturbed instances.

\subsection{Implementation Highlights}
\begin{itemize}
    \item The perturbation strategy employed mirrors the methods used by LIME and SHAP to ensure robustness against common explanation techniques.
    \item The classifier \( \text{is\_OOD} \) is trained using standard off-the-shelf supervised learning algorithms, providing flexibility in its implementation.
    \item Our approach requires only black-box access to the biased classifier \( f \), making it broadly applicable across domains where post hoc explanation techniques are utilized.
\end{itemize}

\section{Methodology, Results, and Discussion}


\begin{figure*}
    \centering
    \includegraphics[width=\textwidth]{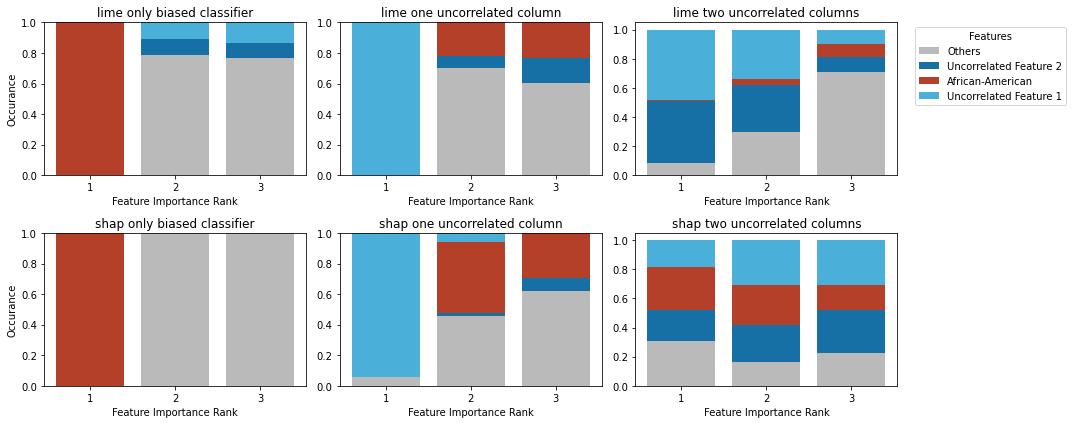} 
    \caption{Replication of top 3 features on the COMPAS dataset}
    \label{fig:wide_figure}
\end{figure*}

\subsection{Replication:}
\cite{slack2020foolinglimeshapadversarial} conducted experiments using multiple datasets, including COMPAS for recidivism risk prediction, Communities and Crime for violent crime prediction, and German Credit for credit scoring. Each dataset includes a sensitive feature: race in COMPAS and Communities and Crime, and gender in German Credit. Following their proposed framework, the authors constructed an adversarial classifier from an existing biased classifier to achieve near-perfect fidelity with respect to the biased classifier. Fidelity in this context refers to the degree to which the adversarial classifier replicates the decisions of the original biased classifier. Predictions were explained by application of SHAP and LIME to the biased and adversarial classifiers. This resulted in finding that while LIME and SHAP accurately identify the sensitive feature as the most important feature in the biased classifier, they deflect the importance from the sensitive attribute to other uncorrelated features when applied to the adversarial classifier, consistently across datasets. This finding is particularly significant, as it demonstrates that LIME and SHAP explanations can be systematically manipulated to obscure biased features using adversarial techniques. In this project, we plan to replicate this experiment, specifically focusing on the COMPAS dataset, to validate the authors' observations regarding the feature importance shifts in adversarial classifiers.

In the prescribed framework to construct an adversarial classifier, \cite{slack2020foolinglimeshapadversarial} used an Out-of-Distribution (OOD) classifier to decide whether or not a given instance is perturbed or from the original distribution. The authors found that the accuracy of the OOD classifier (measured by the F1 metric) significantly impacts the feature attribution by SHAP and LIME on the adversarial classifiers. The more accurate the OOD classifier, the more likely SHAP and LIME provide misleading explanations on the adversarial classifiers. This is an important finding that we will replicate as it forms the basis of the extension of our research project.

For our replication, we focus on two specific figures from  \cite{slack2020foolinglimeshapadversarial}: Figures 6 and 7. Figure 6 represents how well the adversarial models impact LIME and SHAP's capabilities to determine bias. The contrast between the performance of LIME and SHAP against the biased classifier (on the left) as compared to the adversarial classifiers in the middle and on the right is apparent and demonstrates the effectiveness of their adversarial models. Figure 7 shows how effective the adversarial models are at fooling LIME and SHAP with varying accuracies of the OOD classifier. We replicated these figures as they are crucial in determining the effectiveness of our extension.

\begin{figure}
    \centering
    \includegraphics[width=\linewidth]{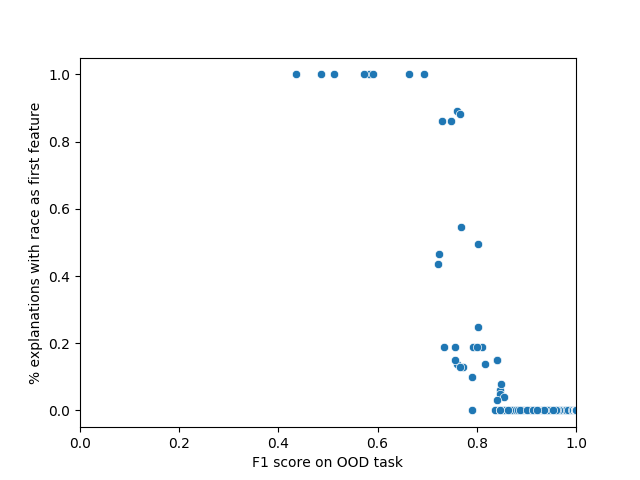}
    \caption{Replicated LIME Sensitivity Analysis}
    \label{fig:enter-label}
\end{figure}
\begin{figure}
    \centering
    \includegraphics[width=\linewidth]{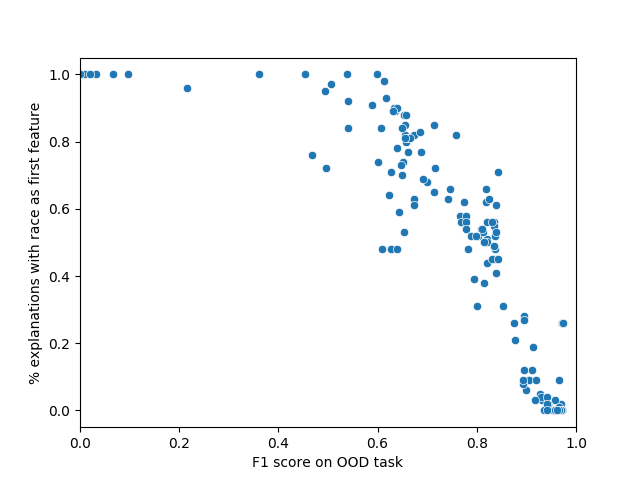}
    \caption{Replicated SHAP Sensitivity Analysis}
    \label{fig:enter-label}
\end{figure}

A majority of the code used to replicate the findings was made publicly available by the authors of the original paper. This included the OOD classifiers for both LIME and SHAP as well as the biased and adversarial classifiers. Specifically, the code that the original paper used to create the graphs in figure 6 was available in whole. Using this code, we created our adversarial classifiers for various accuracies (F1 Scores) and tested how well they fooled LIME and SHAP respectively. In the end, our results were near identical to the paper's (Figures 4 and 5), and we reinforced the results that that the adversarial classifiers fooled LIME and SHAP with varying success over differing accuracies. Particularly, the ability of the adversarial model to fool LIME sharply increased as the F1 score increased above 0.75 while the ability of the model to fool SHAP increased comparatively gradually beginning around an F1 score of 0.5. 

For replicating Figure 7, the code base provided by the original paper's authors supplied a baseline application to the COMPAS dataset. However, replication necessitated slight modifications to the code to calculate and output the values as needed for replication. By using these modified outputs, we were able to recreate the figure with a large degree of similarity (Figure 2). These results reinforce the original paper's proposal that their adversarial classifiers were able to fool LIME and SHAP into thinking a biased classifier was unbiased; this creates the foundation of our research as an extension of their work.



\subsection{Testing Framework:}

\section{Extensions}


\begin{enumerate}
    \item \textbf{Significance:}
    
    Understanding the rationale behind a model's prediction is often just as critical, if not more than, the model's prediction itself. This emphasis on interpretability is at the heart of ExplainableAI (XAI) and fairness. In response, various post-hoc interpretability methods have been developed, with LIME and SHAP emerging as the two of the most prominent techniques. Given that LIME and SHAP are commonly employed to reveal the underlying vulnerabilities of complex models, it is also crucial to examine potential weaknesses in these explanation methods themselves. If adversarial examples can exploit or bypass the interpretative frameworks of LIME and SHAP, the reliability of these methods may be compromised. By demonstrating that adversarial models can manipulate the explanations provided by LIME and SHAP, this research exposes the shortcomings of SHAP and LIME. These findings not only highlight the need to scrutinize these post-hoc explanation methods but also suggest future directions for refining these methods by addressing their shortcomings. Strengthening these methods against adversarial manipulation will enhance their reliability, advancing more robust and trustworthy interpretability in machine learning.


    \item \textbf{Extension(s):}
    For our research, we propose developing SHLIME, a novel ensemble approach that combines LIME and SHAP explanations to create a more robust explanation method against adversarial attacks. Our motivation stems from the paper's key finding that LIME and SHAP exhibit complementary vulnerabilities to Out-of-Distribution (OOD) classifiers: SHAP is partially susceptible to attacks from less accurate OOD classifiers (F1 $\approx$ 0.45) but requires highly accurate classifiers for complete deception, while LIME remains resilient until reaching higher accuracies (F1 $\approx$  0.7) but becomes easily fooled beyond that threshold (F1 $>$ 0.8). See Figure 3 for reference.
    \\
    We hypothesize that by intelligently combining these methods, we can leverage their complementary strengths to create an explanation method that maintains high reliability across a broader range of OOD accuracies. The challenge in combining LIME and SHAP is that their outputs are represented differently — SHAP values lie between 0 and 1, while LIME values can be unbounded and negative. Therefore, whatever method we use to combine them must preserve as much information as possible from both models. The most straightforward approach to doing so was simply multiplying the two values together, using the SHAP values as a scaling factor for LIME's output. Other methods for combining the two models such as taking the weighted average of LIME and SHAP values would result in losing some information from either model, most starkly in cases where the outputs of LIME and SHAP diverged significantly.
    
    We hypothesize that a successful implementation of SHLIME should demonstrate enhanced robustness, remaining unfounded until higher F1 scores similar to LIME, while showing a more gradual degradation in performance with increase in F1 scores similar to SHAP. This behavior contrasts favorably with the earlier and more abrupt vulnerability transitions of individual methods.
    We investigate the performance of BASIC SHLIME, an ensemble method that returns the LIME value multiplied by the SHAP value for each feature. 

    \begin{figure}[t]
        \centering
        \includegraphics[width=\linewidth]{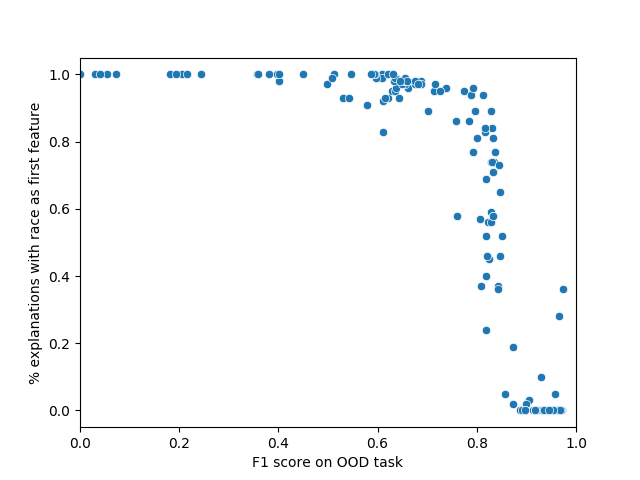}
        \caption{BASIC SHLIME Sensitivity Analysis}
        \label{fig:enter-label}
    \end{figure}
\end{enumerate}

\subsection{Societal Impact}

The societal impact of the paper is profound, as it directly addresses the integrity and fairness of machine learning models deployed in critical sectors such as healthcare, finance, and criminal justice. Black-box models, which are increasingly used to make high-stakes decisions in these areas, often operate without sufficient transparency, raising concerns about biases, fairness, and accountability. This research highlights vulnerabilities in popular post hoc explanation methods like LIME and SHAP, revealing how easily these models can be manipulated, leading to potentially harmful consequences when they are used to inform decisions about individuals' lives. By uncovering these weaknesses, the paper emphasizes the need for more robust explanation techniques, pushing for a more rigorous approach to model testing and evaluation. The ultimate goal is to build models that are not only accurate but also equitable and transparent, ensuring that they do not perpetuate existing biases or make decisions based on harmful stereotypes. By advocating for the development of higher-quality testing methods, this research aims to improve decision-making processes in sensitive fields, reducing the risks of unjust outcomes and enhancing public trust in AI systems.

\section{Conclusions}

As shown in Figure 5, our experimental results demonstrate that this simple yet effective approach successfully improves upon both methods. BASIC SHLIME maintains strong explanation accuracy (consistently above 0.8) for a wider range of F1 scores (similar to LIME), only beginning to show significant degradation around F1 $\approx$ 0.75. Even as the classifier accuracy increases beyond this threshold, SHLIME exhibits a more gradual decay in performance (similar to SHAP). By leveraging the complementary characteristics of SHAP and LIME, BASIC SHLIME provides more reliable explanations than SHAP and LIME even as adversarial models become more sophisticated.

\subsection{Future Work}
The natural path to take after this research would be to create a proper OOD classifier for our potential SHLIME models. While our results using both the LIME and SHAP OOD classifiers are promising, the strength of our results is limited by the inability to truly compare its robustness on a fair scale in comparison to LIME and SHAP individually. We initially intended to address this within this research project, but the scale and complexity of the issue quickly outgrew our scope. Solutions for this problem would vary immensely from prior research due to our methods combining both LIME and SHAP, making it more suitable as an area for future study. Moreover, each different possible option for SHLIME would require a completely different OOD classifier.

Secondly, the focus of this research was mainly on the robustness of our SHLIME implementation, specifically to these kinds of adversarial attack. However, it is still important to consider how well it performs in its intended interpretability task compared to LIME and SHAP. In future research, it would be useful to see how SHLIME performs in comparison to both LIME and SHAP on a wide array of datasets. Ideally, SHLIME would perform comparably or even better than SHAP, but by introducing LIME to increase robustness, there may potentially be some performance decrease.

Another avenue for future work is to explore alternative methods of combining the outputs of LIME and SHAP. One such approach is the mixture of experts method.: an approach that divides a model into several subnetworks ( or ``experts''), each specializing in a different subset of the input data. For the purpose of this project, one set of experts can specialize in LIME while the other specializes in SHAP. The model result will then be the combined output of the LIME experts and the SHAP experts.



\bibliography{aaai24}

\begin{thebibliography}{6}
\providecommand{\natexlab}[1]{#1}

\bibitem[{Dombrowski et~al.(2019)Dombrowski, Alber, Anders, Ackermann, Müller, and Kessel}]{dombrowski2019explanationsmanipulatedgeometryblame}
Dombrowski, A.-K.; Alber, M.; Anders, C.~J.; Ackermann, M.; Müller, K.-R.; and Kessel, P. 2019.
\newblock Explanations can be manipulated and geometry is to blame.
\newblock arXiv:1906.07983.

\bibitem[{Ghorbani, Abid, and Zou(2018)}]{ghorbani2018interpretationneuralnetworksfragile}
Ghorbani, A.; Abid, A.; and Zou, J. 2018.
\newblock Interpretation of Neural Networks is Fragile.
\newblock arXiv:1710.10547.

\bibitem[{Lipton(2017)}]{lipton2017mythosmodelinterpretability}
Lipton, Z.~C. 2017.
\newblock The Mythos of Model Interpretability.
\newblock arXiv:1606.03490.

\bibitem[{Lundberg and Lee(2017)}]{shap2017}
Lundberg, S.~M.; and Lee, S. 2017.
\newblock A unified approach to interpreting model predictions.
\newblock \emph{CoRR}, abs/1705.07874.

\bibitem[{Ribeiro, Singh, and Guestrin(2016)}]{lime2016}
Ribeiro, M.~T.; Singh, S.; and Guestrin, C. 2016.
\newblock "Why Should I Trust You?": Explaining the Predictions of Any Classifier.
\newblock In \emph{Proceedings of the 22nd ACM SIGKDD International Conference on Knowledge Discovery and Data Mining}, KDD '16, 1135–1144. New York, NY, USA: Association for Computing Machinery.
\newblock ISBN 9781450342322.

\bibitem[{Slack et~al.(2020)Slack, Hilgard, Jia, Singh, and Lakkaraju}]{slack2020foolinglimeshapadversarial}
Slack, D.; Hilgard, S.; Jia, E.; Singh, S.; and Lakkaraju, H. 2020.
\newblock Fooling LIME and SHAP: Adversarial Attacks on Post hoc Explanation Methods.
\newblock arXiv:1911.02508.

\end{thebibliography}
\section{Appendix}
\vspace{-5em}
Access to our experiment repository can be found at: https://github.com/eduguet/shlime
\vspace{-5em}
\subsection{Reference Figures from Slack et al.}
\vspace{-5em}
\begin{figure}[H]
    \includegraphics[width=0.9\linewidth]{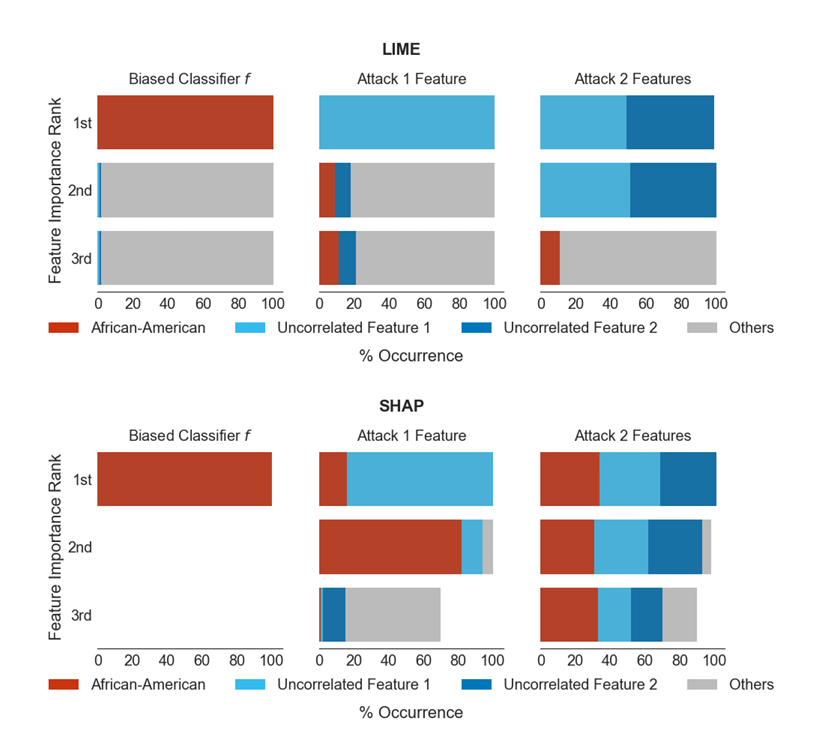}
    \caption{COMPAS: \% of data points for which each feature shows up in top 3 (according to LIME and SHAP’s ranking of feature importance) for different classifiers}
    \label{fig:lime-shap fooled}
    \hspace{\fill}
\end{figure}
\begin{figure}[H]
    \includegraphics[width=0.9\linewidth]{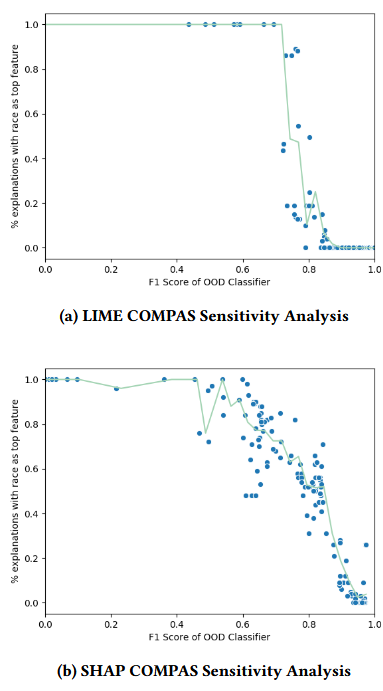}
    \caption{Reference graph of LIME and SHAP being fooled by biased OOD classifiers}
    \label{fig:feature ranking}
    
\end{figure}

\end{document}